\documentclass[conference]{IEEEtran}
\pdfoutput=1    

\usepackage{xcolor}
\definecolor{wong-black}        {HTML}{000000}
\definecolor{wong-lightorange}  {HTML}{E69F00}
\definecolor{wong-lightblue}    {HTML}{56B4E9}
\definecolor{wong-green}        {HTML}{009E73}
\definecolor{wong-yellow}       {HTML}{F0E442}
\definecolor{wong-darkblue}     {HTML}{0072B2}
\definecolor{wong-darkorange}   {HTML}{D55E00}
\definecolor{wong-pink}         {HTML}{CC79A7}

\usepackage{url}

\usepackage{hyperref} 
\hypersetup{
    colorlinks=true,
    citecolor=wong-green,
    linkcolor=wong-darkblue,
    filecolor=wong-pink,      
    urlcolor=wong-black
    }

\usepackage{cite}
\usepackage{amsmath,amssymb,amsfonts}
\usepackage{algorithmic}
\usepackage{graphicx}
\usepackage{subcaption}
\usepackage{textcomp}
\usepackage[nolist, nohyperlinks, printonlyused]{acronym} 

\usepackage{booktabs}

\def\BibTeX{{\rm B\kern-.05em{\sc i\kern-.025em b}\kern-.08em
    T\kern-.1667em\lower.7ex\hbox{E}\kern-.125emX}}
    
\newcommand\nnfootnote[1]{  
  \begin{NoHyper}
  \renewcommand\thefootnote{}\footnote{#1}%
  \addtocounter{footnote}{-1}%
  \end{NoHyper}
}

\begin{document}
\bstctlcite{IEEEexample:BSTcontrol}


\title{
Conditioning Latent-Space Clusters\\for Real-World Anomaly Classification
}


\author{\IEEEauthorblockN{Daniel Bogdoll\IEEEauthorrefmark{2}\IEEEauthorrefmark{3}\textsuperscript{\textasteriskcentered}, 
Svetlana Pavlitska\IEEEauthorrefmark{2}\IEEEauthorrefmark{3}\textsuperscript{\textasteriskcentered},
Simon Klaus\IEEEauthorrefmark{3}\textsuperscript{\textasteriskcentered},
and J. Marius Zöllner\IEEEauthorrefmark{2}\IEEEauthorrefmark{3}}

\IEEEauthorblockA{\IEEEauthorrefmark{2}FZI Research Center for Information Technology, Germany\\
bogdoll@fzi.de}
\IEEEauthorblockA{\IEEEauthorrefmark{3}Karlsruhe Institute of Technology, Germany\\}}

\maketitle

\nnfootnote{\textasteriskcentered~These authors contributed equally}


\begin{acronym}
    \acro{ml}[ML]{Machine Learning}
	\acro{cnn}[CNN]{Convolutional Neural Network}
	\acro{dl}[DL]{Deep Learning}
	\acro{ad}[AD]{Autonomous Driving}
\end{acronym}


\begin{abstract}
Anomalies in the domain of autonomous driving are a major hindrance to the large-scale deployment of autonomous vehicles. In this work, we focus on high-resolution camera data from urban scenes that include anomalies of various types and sizes. Based on a Variational Autoencoder, we condition its latent space to classify samples as either normal data or anomalies. In order to emphasize especially small anomalies, we perform experiments where we provide the VAE with a discrepancy map as an additional input, evaluating its impact on the detection performance. Our method separates normal data and anomalies into isolated clusters while still reconstructing high-quality images, leading to meaningful latent representations.
\end{abstract}




\section{Introduction}
\label{sec:introduction}

Developing autonomous vehicles and deploying them to large Operational Design Domains (ODD) poses a significant challenge, especially with respect to the long tail of unexpected or unfamiliar objects. Although perception systems of autonomous vehicles are able to detect known classes reasonably well nowadays, they still need to be aware of situations where they encounter the unknown. As deep neural networks (DNN) tend to predict false positives with high uncertainty, both false negatives and false positives are of interest. While often lidar sensors are used in production systems, here we focus on a purely camera-based setup, as those introduce their own set of issues. We focus on detecting anomalies in high-resolution road images from mostly urban scenes based on the latent space of a Variational Autoencoder (VAE). Utilizing primarily normal but also abnormal data during training, the data is being fit on two prior distributions. This way, the VAE is conditioned to build two separate clusters in the latent space, one for normal samples and one for anomalies. During test time, distance measures can be used to detect anomalies. We used multiple datasets to define normality and anomalies during training and evaluation. The work is structured as follows: In Section~\ref{sec:related_work}, we introduce related work from the field of anomaly detection, with a focus on Variational Autoencoders. In Section~\ref{sec:method}, we introduce our approach, including our VAE architecture. In Section~\ref{sec:evaluation}, we highlight our experimental setup and demonstrate our results. Finally, we conclude this work in Section~\ref{sec:conclusion}. More information can be found in~\cite{Klaus_Anomaly_2022_BA}. 

\begin{figure}
\includegraphics[width=\columnwidth]{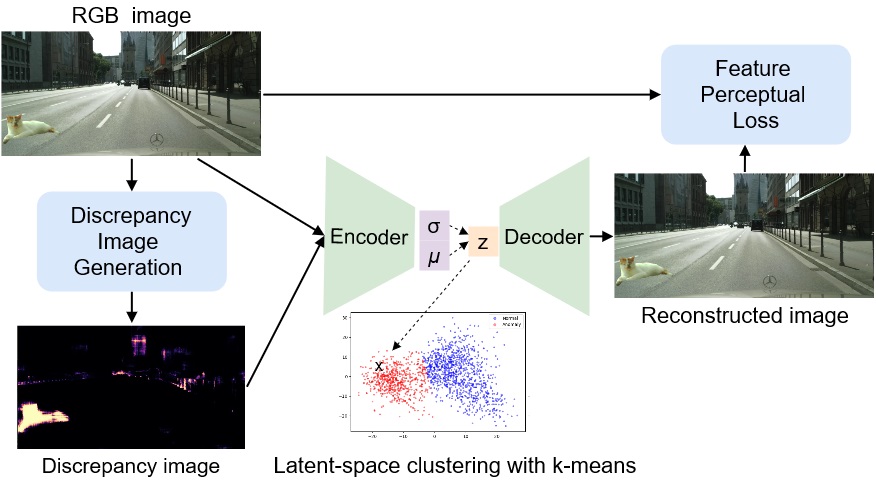}
\centering
\caption{Our VAE-based method for real-world anomaly classification, which separates normal and abnormal data in its latent space. Discrepancy images as additional inputs also emphasize small unknown objects, here \textit{a cat}.}
\label{fig:compare_sensors}
\end{figure}

\section{Related Work}
\label{sec:related_work}

In autonomous driving, detecting anomalies is of utmost importance to scale existing systems, which operate in small Operational Design Domains, as infrequent events occur more often for a growing vehicle fleet that utilizes the same software system~\cite{houben2021inspect}.
Based on common corner case systematizations~\cite{breitenstein_systematization_2020, heidecker_application-driven_2021, Bogdoll_Description_2021_ICCV}, we are especially interested in the \textit{object} and \textit{scene} levels, which describe unknown objects or, more generally, unexpected patterns in an input sample. In this section, we introduce current approaches for such detections, also known as outliers or out-of-distribution (OOD) samples.

As one of the early approaches, Ruff et al.~\cite{ruff_deep_2018} trained a deep neural network in order to compute a hypersphere representing an approximation of normality. A binary classification can be computed based on the distance of new samples to the hypersphere. However, in the real world, normality is represented in the form of many “heterogeneous semantic labels”~\cite{park_what_2021}, which leads to a weak decision boundary. Hendrycks et al. proposed an “outlier exposure” objective~\cite{hendrycks_deep_2018}, utilizing curated anomaly data for training, leading to a more uniform softmax distribution for anomalies, which was later adopted by Papadopoulos et al.~\cite{papadopoulos_outlier_2021}. Some approaches utilize the softmax confidence in order to detect anomalies~\cite{hendrycks_baseline_2022, liang_enhancing_2018}. However, this can lead to false positives since the softmax activation function is sensitive to changes in the input. Thus, Liu et al. proposed an energy-based approach~\cite{liu_energy-based_2020}, showing an improved alignment to the density of the input samples.

Based on a GAN, Nitsch et al. generated virtual anomalies for an improved decision boundary~\cite{nitsch_out--distribution_2021}. Similarly, Grcic et al. used normalizing flows for the same task~\cite{grcic_dense_2021}, outperforming most other methods~\cite{bogdoll_anomaly_2022}. Going one step further, Du et al. introduced Virtual Outlier Synthesis (VOS), which generates synthetic anomalies in the latent space~\cite{du_vos_2022}. In the field of semantic segmentation, Cen et al. included unknown objects in their class list, leading to an open-world segmentation approach based on Euclidean distances between feature vectors~\cite{cen_deep_2021}. Di Biase integrated model uncertainty into their system in order to reduce wrong classifications~\cite{di_biase_pixel-wise_2021}.

As we have shown, there exist many different methods to detect anomalies. Since our work is based on VAEs, we now provide a detailed overview of methods based on encodings.

\subsection{Encoding-based Anomaly Detection}

Breitenstein et al. have categorized anomaly detection in the domain of autonomous driving into five different categories: Reconstruction, Prediction, Generative, Confidence Score, Feature Extraction~\cite{breitenstein2021corner}. In this work, we examine the properties of Variational Autoencoders in order to classify image samples as anomalies.  VAEs can be utilized for dense anomaly detections, which fall under either the \textit{Reconstruction} or \textit{Generative category}, and anomaly classifications, which can be categorized as \textit{Feature Extraction}. Methods from the first category are based on the assumption that the utilized training data defines normality, leading to failed reconstructions given samples that include parts that were not included in the training data. Methods from the second category, which take a look at the latent space, assume that latent representations from normal and OOD samples have a sufficient difference.

\textbf{Reconstructive and Generative.} A well-trained VAE will reconstruct unseen anomalies~\cite{Bogdoll_Compressing_2021_NeurIPS}, which is why specialized methods were developed for anomaly detection. Utilizing a Generative Adversarial Network (GAN), in which both the generator and the discriminator are implemented as AE, Vu et al. designed a network where the discriminator learns to reconstruct normal data while failing to do so when presented with OOD data~\cite{vu_anomaly_2019}. Utilizing the reconstruction probability, An et al. go beyond utilizing the direct reconstruction error, which is incapable of incorporating high-level structures~\cite{an_variational_2015}. Among others, Munjal et al. introduced an adversarial loss in order to address this issue~\cite{munjal_implicit_2020}. However, this method is not effective for high-complexity scenes. On a similar note, Somepalli et al. proposed to minimize the Wasserstein distance and include a latent space regularization, which led to better reconstructions of normal data and worse ones for anomalies. On the other hand, Bolte et al.~\cite{bolte_towards_2019} proposed a multi-stage approach, where an Autoencoder was used for image prediction. Based on this, an engineered approach followed, which included prediction errors, pixel classification, and distance weighting. Similarly, Amini et al.~\cite{amini_variational_2018} proposed a pixel-wise uncertainty for reconstructions with a VAE. It is also possible to combine both methods which are based on reconstructions and feature extraction. Abati et al.~\cite{abati_latent_2019} utilize reconstruction error and latent features with a low log-likelihood in order to detect anomalies. Similarly, Wang et al.~\cite{wang_image_2020} use a discrete latent space, where a model learns the distribution. Reconstructions are then based upon a re-sampled latent space and used for anomaly detection. Park et al.~\cite{park_learning_2020} proposed a memory module where prototypes of normality are stored. Later, a reconstruction-based approach, using these memory items, is used for anomaly detection. 

\begin{figure}[t]
    \centering
    \includegraphics[width=\columnwidth]{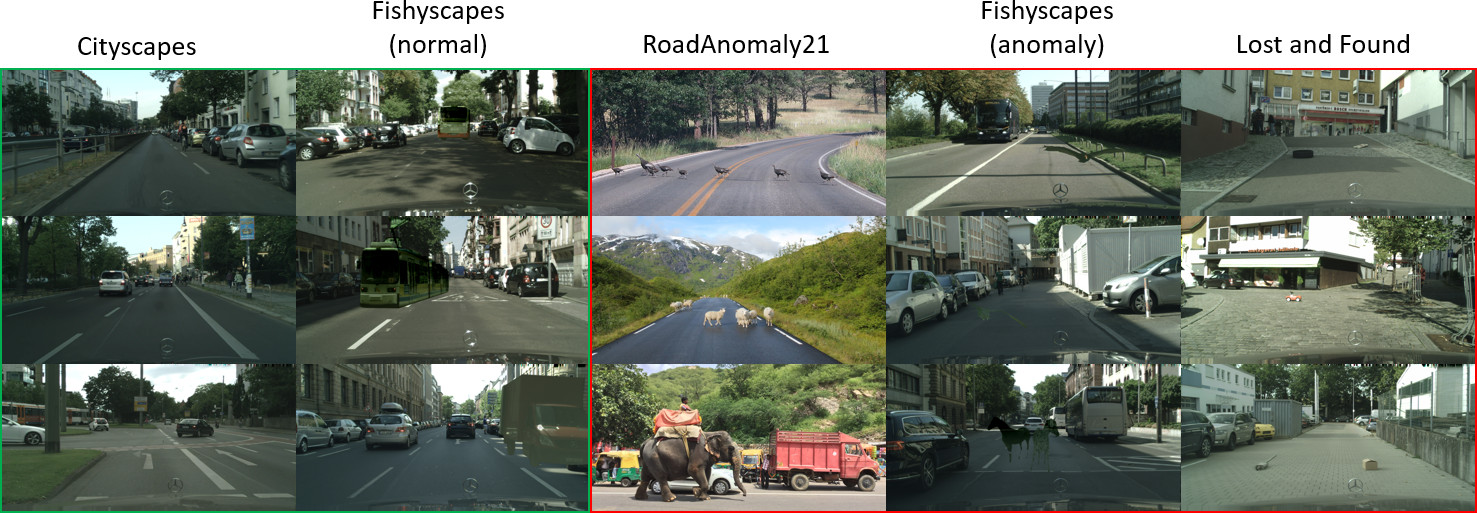}
    \caption{We used the Cityscapes~\cite{cordts_cityscapes_2016} and Fishyscapes~\cite{blum_fishyscapes_2021} (normal) datasets as normality (left) and the RoadAnomaly21~\cite{chan_segmentmeifyoucan_2021}, Fishyscapes (anomalies), and Lost and Found~\cite{pinggera_lost_2016} datasets with anomalies (right). Reprinted from~\cite{Klaus_Anomaly_2022_BA}.}
    \label{fig:datasets}
\end{figure}

\textbf{Feature Extraction.} As some previous techniques already utilized feature extraction partly, this section highlights works focusing on this technique. Wurst et al.~\cite{wurst_novelty_2021} utilized a triplet-based Autoencoder, enforcing similarity in the latent space, to detect unusual traffic scenes. Similarly, Harmening et al.~\cite{harmening_deep_2020} uses clusters in the latent space generated by an Autoencoder to detect novel scenarios. For more complex data, Sundar et al.~\cite{sundar_out--distribution_2020} developed a method to divide datasets into smaller subsets. Based on these, multiple VAEs are trained to generate the latent space. For detection, they utilize all trained VAEs and detect high sensitivity. Akcay et al.~\cite{akcay_ganomaly_2019} compared latent representations of image reconstructions and the original input to detect anomalies. Chalapathy at al.~\cite{chalapathy_anomaly_2019} utilize a one-class classification objective based on features learned by a VAE, where they focus on generating features that are designed for the task of anomaly detection~\cite{erfani_high-dimensional_2016}. Park et al.~\cite{park_interpreting_2020} utilize rate-distortion theory in order to compute anomaly scores, only using the encoding part of a VAE. The work of Liu et al.~\cite{liu_towards_2020} is based on attention maps for every element of the latent vector, where they compute differences to the learned normality, leading to an attention map that highlights anomalies in an image. Finally, Dilokthanakul et al.~\cite{dilokthanakul_deep_2017} proposed a VAE which uses a mixture of Gaussians as prior, assuming multiple distributions in the training data, which led to a better separation of classes in the latent space.

While many of the presented approaches work with simple datasets, in our work, we are interested in high-resolution images~\cite{Bogdoll_Addatasets_2022_VEHITS, Bogdoll_Impact_2023_ICCRE} with anomalies in urban road scenarios~\cite{Bogdoll_Perception_2023_IV}. Here, the challenge arises that anomalies often only occupy small regions of an image, which makes classification harder, as normality is represented by highly complex training data. Our approach evaluates whether an auxiliary input that highlights even small anomalies in the image space, combined with a conditioned latent space, allows for the classification of anomalies in such datasets.

\section{Method}
\label{sec:method}


\begin{figure}[t]
	\centering
	\begin{subfigure}[b]{0.69\columnwidth}
		\centering
		\includegraphics[height=2.4cm]{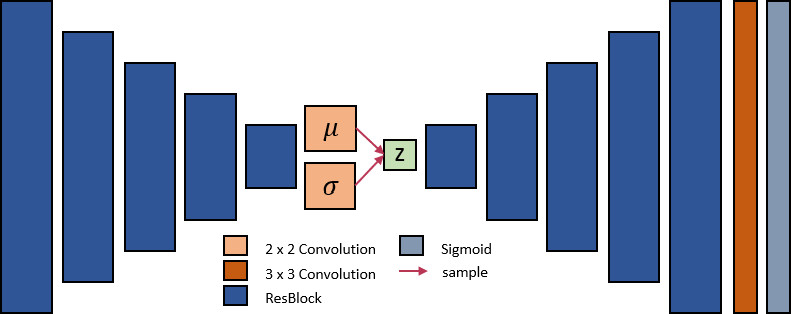}
		\caption{Overall architecture of the VAE.}
		\label{fig:vae_architecture}
	\end{subfigure}
	\begin{subfigure}[b]{0.29\columnwidth}
		\centering
		\includegraphics[height=2.4cm]{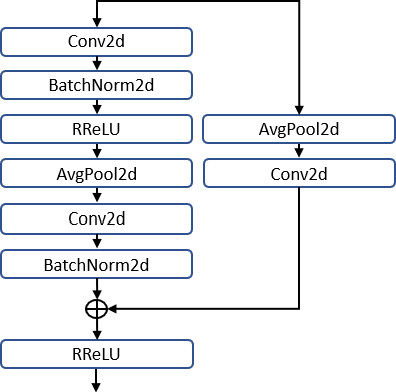}
		\caption{ResBlock}
		\label{fig:resblock}
	\end{subfigure}
	\caption{Overall architecture of the deployed VAE (left) and the components of the ResBlock (right). Adapted from~\cite{Klaus_Anomaly_2022_BA}}
	\label{fig:combined}
\end{figure}


In the following, we describe our anomaly detection method, which conditions the latent space of a VAE to enforce separations of clusters corresponding to anomalous and normal data.

\subsection{VAE Architecture}

We use a conditioned latent space variational autoencoder (CL-VAE)~\cite{norlander_latent_2019}. Our architecture roughly follows Hou et al.~\cite{hou_deep_2017}. However, we utilize residual blocks, as shown in Figure~\ref{fig:vae_architecture}\footnote{Implementation inspired by \href{https://github.com/LukeDitria/CNN-VAE}{LukeDitria/CNN-VAE}}, as those are easier to train. Furthermore, we inserted an additional pooling layer and adjusted the residual block by replacing the exponential linear unit (ELU) activation function with the randomized leaky rectified linear units (RRelu) as proposed by Xu et al.~\cite{xu_empirical_2015}, see Figure \ref{fig:resblock}. The VAE was trained with the CL-VAE ELBO loss. We performed experiments with two auxiliary losses to enforce the separation of normal and anomalous data. First, the distance loss $\mathcal{L}_{distance}$ aims at maximizing the distance between the two means to push the clusters away from each other:
\begin{align}
    \label{eq:dist_loss}
    \mathcal{L}_{distance} = - \| \mu_1, \mu_2 \|_1 = - |\mu_1 - \mu_2|
\end{align}

Second, we use $\mathcal{L}_{i}$ proposed by Yang et al.~\cite{yang_towards_2017} to maximize the distance between each data point and the cluster mean given by the k-means algorithm: $\mathcal{L}_{i} = (\mu_i - z_i)^2$. The overall \textit{cluster loss} is then given by:
\begin{align}
    \mathcal{L}_{cluster} = \frac{1}{n} \sum_{i=1}^{n} \mathcal{L}_i    
    \label{eq:cluster_loss}
\end{align}

We also incorporate the feature perceptual loss~\cite{hou_deep_2017} using a pre-trained backbone to enforce reconstruction quality. This concept ensures meaningful latent representations of the input samples, which can be used for downstream tasks. Discrepancy images passed as the fourth channel were not considered as the backbone was pre-trained on RGB images.


\subsection{Generation of Discrepancy Images}
We use discrepancy images, highlighting areas with anomalous objects' locations, as an additional input to a VAE as a fourth channel. Following the method proposed by Lis et al.~\cite{lis_detecting_2019}, we first create a semantic segmentation prediction for a given image. A GAN then tries to recreate the original image from this semantic segmentation image. Finally, the discrepancy network is used to generate the discrepancy image by comparing this recreated image to its counterpart. 
The discrepancy network comprises three streams: a pre-trained CNN extract features from an original and a resynthesized image, and a custom CNN extracts features from a semantic segmentation map. The extracted features pass through a decoder which outputs the resulting discrepancy image.

\begin{figure}[t]
	\centering
	\includegraphics[width=\columnwidth]{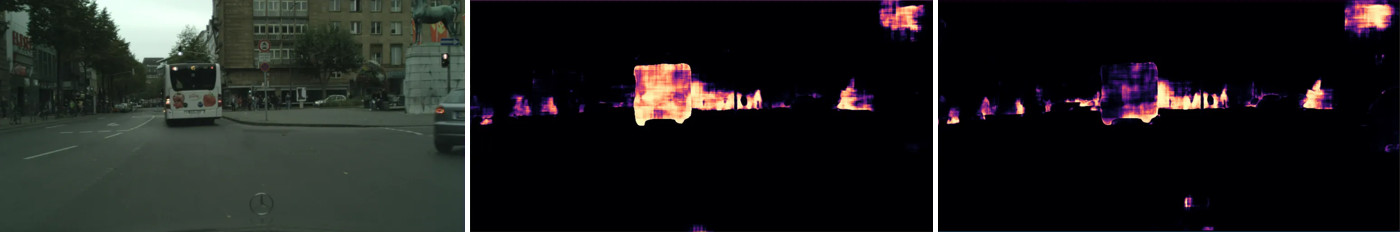}
	\caption{Discrepancy images for a Cityscapes image containing an object of the rare but normal class \textit{bus}. The original approach by Lis et al.~\cite{lis_detecting_2019} (middle) leads to higher anomaly scores. The proposed frequency-based approach (right) leads to lower anomaly scores. Reprinted from~\cite{Klaus_Anomaly_2022_BA}.}
	\label{fig:discrepancy_bus}
\end{figure}

In the original approach by Lis et al.~\cite{lis_detecting_2019}, the discrepancy detector is trained on the dataset of normal data with altered labels. In particular, labels of some randomly selected objects are replaced with random class labels, thus creating synthetic anomalies. However, due to natural class imbalance, the model learned to classify objects of rare classes as anomalies because randomly choosing a replacement class makes them occur more frequently as an anomaly replacement class than a normal class. To mitigate this issue, we propose the \textit{frequence-based label replacement} as shown in Figure~\ref{fig:discrepancy_bus}. To create a synthetic anomaly dataset for training, rare classes, i.e., those which occur less frequently in a dataset of normal data, are chosen as frequently as a replacement as common ones.


\section{Evaluation}
\label{sec:evaluation}

In the following, we describe the evaluation of our anomaly detection method. First, we provide details on our experimental setup, followed by several analyses. 

\subsection{Experimental Setting}
\textbf{Training Data:} We utilized three datasets to train the VAE. \textit{Cityscapes}~\cite{cordts_cityscapes_2016} was used to to represent normal data and both \textit{LostAndFound}~\cite{pinggera_lost_2016} and \textit{RoadAnomaly21} from the \textit{SegmentMeIfYouCan} benchmark~\cite{chan_segmentmeifyoucan_2021} were used to represent anomalous data. Samples from these datasets can be found in Figure~\ref{fig:datasets}. For Cityscapes, we used the pre-defined train-val-test split. The LostAndFound dataset was filtered as follows: We deleted images with less than 3,000 anomalous pixels per image and images containing children, as those are considered normal in Cityscapes. We have selected only a few images with different anomalies from each scene to avoid overfitting. The resulting filtered dataset thus contained 172 train, 99 validation, and 64 test images. Finally, all 110 images from the RoadAnomaly21 dataset were split according to the 70:20:10 rule. For training, all images were downsampled to $256\times256$.

\textbf{Test Data:} For evaluation, the test data from LostAndFound and RoadAnomaly21 datasets were used, which were split as described above. We also used \textit{FS Static} images from the \textit{Fishyscapes} dataset~\cite{blum_fishyscapes_2021}. Because of the small dataset size, it was only used at the test stage. Just 30 images are publicly available, 10 with normal and 20 with anomalous data.

\textbf{Models and Training:} Following the approach proposed by Lis et al.~\cite{lis_detecting_2019}, we used a pre-trained PSPNet~\cite{zhao_pyramid_2017} with a pre-trained ResNet backbone~\cite{he_deep_2016} to predict semantic segmentation masks for input images and a pre-trained pix2pixHD model~\cite{wang_high-resolution_2018} for image resynthesis. The discrepancy module included a pre-trained VGG~\cite{Simonyan15} for feature extraction. The VAE was trained for 100 epochs using the ADAM optimizer~\cite{kingma_ADAM_2017} with a learning rate of $1\text{e-}4$
 and a batch size of 12. The learning rate decreased linearly during training. All trainings were performed on an Nvidia GeForce RTX 3090.

\begin{figure}[t]
    \centering
    \includegraphics[width=0.9\linewidth]{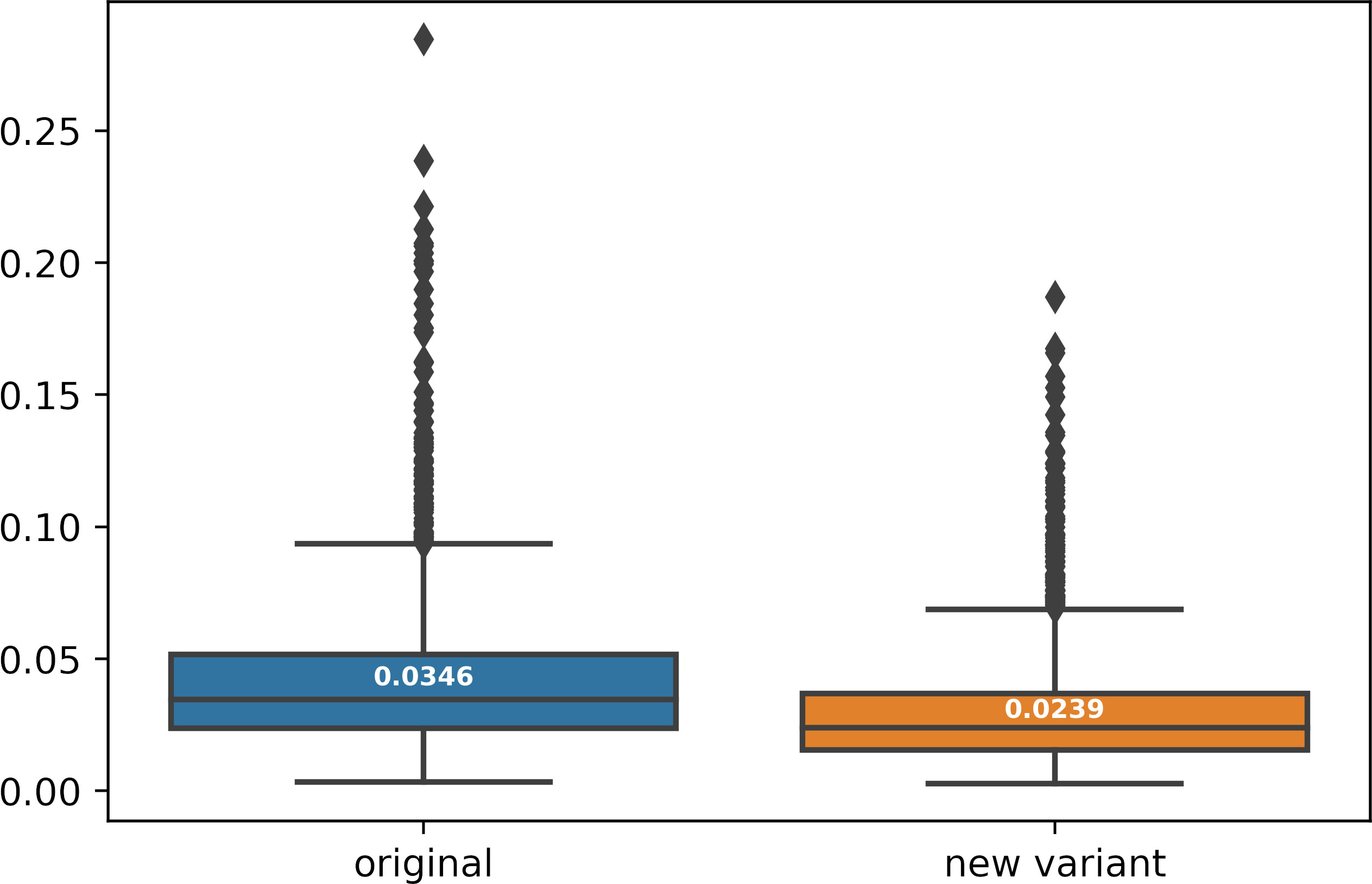}
    \caption{Distribution of mean anomaly scores in the discrepancy maps generated for the Cityscapes test set, comparing the original approach by Lis et al.~\cite{lis_detecting_2019} (blue) to our frequency-based label replacement (orange). Reprinted from~\cite{Klaus_Anomaly_2022_BA}.}
    \label{fig:box_plot}
\end{figure}

\begin{figure}[t]
    \centering
    \begin{subfigure}[b]{0.49\linewidth}
    \centering
    \includegraphics[width=\textwidth]{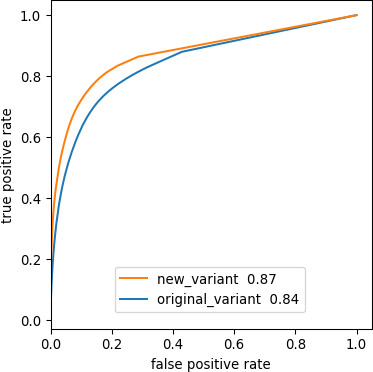}
    \end{subfigure}
    \begin{subfigure}[b]{0.49\linewidth}
    \centering
    \includegraphics[width=\textwidth]{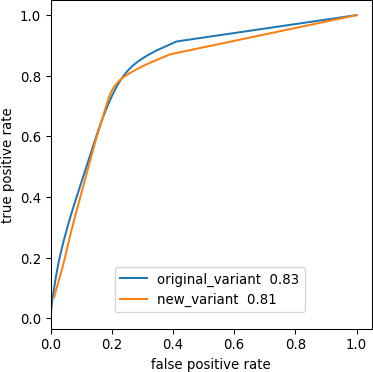}
    \end{subfigure}
    \caption{ROC curves for anomaly detection using the discrepancy maps generated with the proposed frequency-based label replacement: LostAndFound (left) and RoadAnomaly (right) test data. Reprinted from~\cite{Klaus_Anomaly_2022_BA}.}
    \label{fig:disc_roc}
\end{figure}

\subsection{Impact of Frequency-based Label Replacement}
In our discrepancy module, we used Cityscapes as a normal dataset where no anomalies should appear. We have analyzed the average pixel-wise anomaly score in generated grayscale discrepancy images, where 0 corresponds to normal and 1 to anomalous data. Ideally, all discrepancy scores should be zero, as no anomalies exist in the data. As Figure~\ref{fig:box_plot} demonstrates, the average pixel value in discrepancy maps is lower for the proposed frequency-based label replacement variant than the random-class approach proposed by Lis et al.~\cite{lis_detecting_2019}. Furthermore, a visual comparison of the resulting discrepancy images as shown in Figure~\ref{fig:discrepancy_bus} confirms that our frequency-based class selection results in lower anomaly scores for normal classes. Furthermore, we evaluated the impact of the frequency-based class selection on LostAndFound and RoadAnomaly using the anomaly detector from Lis et al.~\cite{lis_detecting_2019}. Figure~\ref{fig:disc_roc} shows that our approach leads to improved classifications for RoadAnomaly dataset but worse results for LostAndFound.



\subsection{VAE Reconstruction Performance}
\begin{figure}[h!]
    \centering
    \includegraphics[width=1\linewidth]{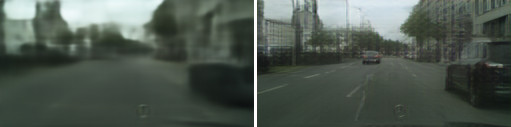}
    \caption{Image reconstruction by a VAE with the latent space size of $512\times4\times4$ and a $\beta=0.01$  trained with (right) and without (left) the feature perceptual loss. Reprinted from~\cite{Klaus_Anomaly_2022_BA}.}
    \label{fig:feat_vs_nofeat}
\end{figure}

\begin{figure*}[t]
    \centering
    \includegraphics[width=1\textwidth]{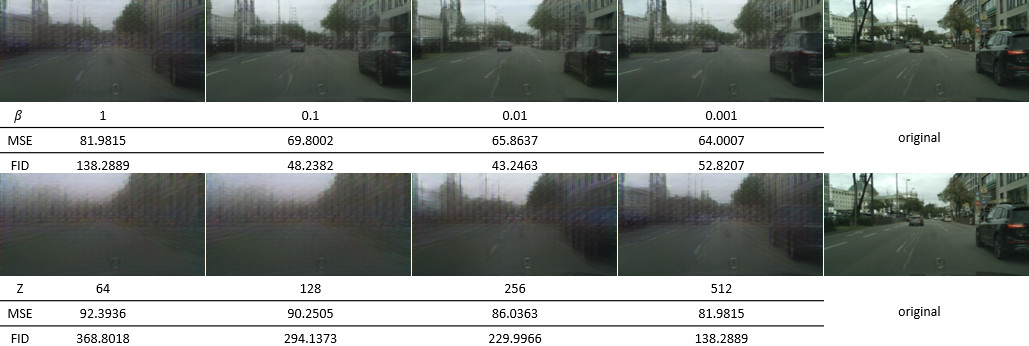}
    \caption{Image reconstructions for different $\beta$ values (top) and latent map sizes (bottom) of a VAE with the latent feature map of size $z\times4\times4$. Average FID and MSE values were measured on the Cityscapes test dataset. Adapted from~\cite{Klaus_Anomaly_2022_BA}.}
    \label{fig:kl_weight_ls_size}
\end{figure*}

We evaluated the impact of two hyperparameters on the reconstruction performance: The size of the latent space and the $\beta$ parameter of the KL divergence. We used the Fréchet Inception Distance (FID)~\cite{heusel_gans_2017} to measure the quality of the reconstructions. Figure~\ref{fig:kl_weight_ls_size} shows that both a larger latent space and smaller $\beta$ lead to more accurate reconstructions. Finally, we evaluated the impact of the feature perceptual loss on the reconstruction quality. Figure~\ref{fig:feat_vs_nofeat} shows that using the feature perceptual loss results in less blurry images.

\subsection{Impact of Discrepency Image}
To evaluate the effect of the discrepancy maps, we first calculated the mean pixel scores for both normal and abnormal data. We found that the score is much higher for images including anomalies than those without. However, an ablation study without the input revealed that the discrepancy map had little effect on the structure of the latent space, especially high-dimensional latent states.

\subsection{Anomaly Classification via Clustering}

\begin{figure}[t]
    \centering

    \begin{subfigure}[b]{0.49\linewidth}
        \centering
        \captionsetup{justification=centering}
        \includegraphics[width=\textwidth]{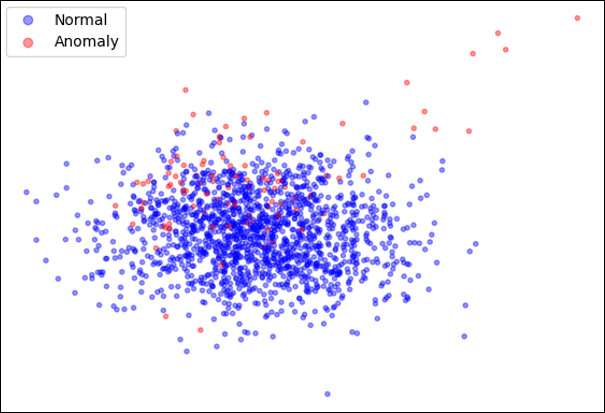}
        \caption{Groundtruth: Latent space size of  $64\times4\times4$}
    \end{subfigure}
    \begin{subfigure}[b]{0.49\linewidth}
        \centering
        \captionsetup{justification=centering}
        \includegraphics[width=\textwidth]{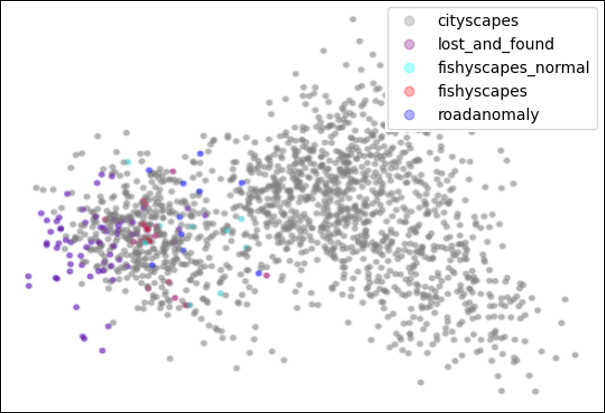}
        \caption{Utilized datasets for VAE with $\beta=0.01$}
    \end{subfigure}
    
    \begin{subfigure}[b]{0.49\linewidth}
        \centering
        \captionsetup{justification=centering}
        \includegraphics[width=\textwidth]{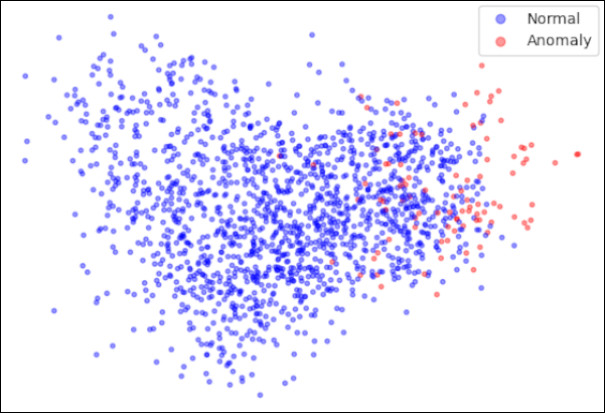}
        \caption{Groundtruth: Latent space size of $512\times4\times4$.}
    \end{subfigure}
    \begin{subfigure}[b]{0.49\linewidth}
        \centering
        \captionsetup{justification=centering}
        \includegraphics[width=\textwidth]{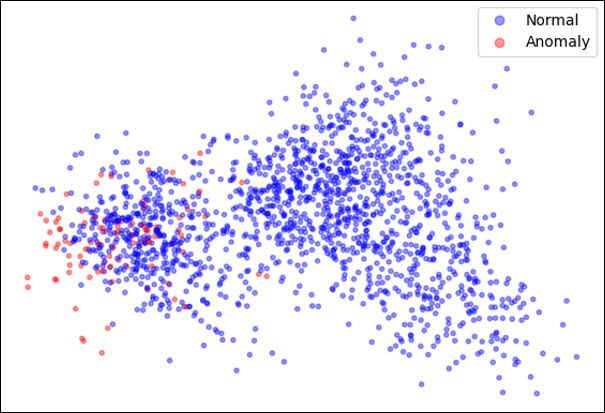} 
        \caption{Groundtruth: VAE with $\beta=0.01$}
    \end{subfigure}
    
    \begin{subfigure}[b]{0.49\linewidth}
        \centering
        \captionsetup{justification=centering}
        \includegraphics[width=\textwidth]{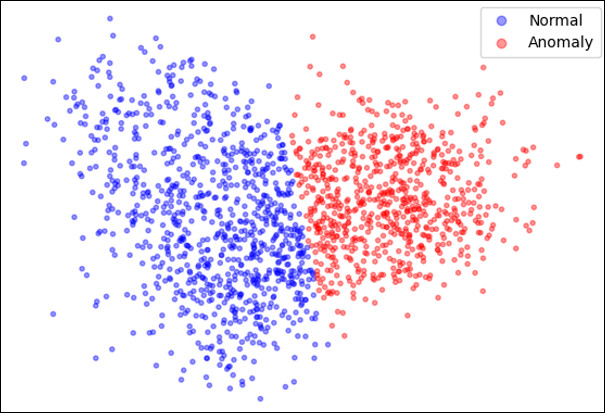}
        \caption{Clustering: Latent space of size $512\times4\times4$.}
    \end{subfigure}
    \begin{subfigure}[b]{0.49\linewidth}
        \centering
        \captionsetup{justification=centering}
        \includegraphics[width=\textwidth]{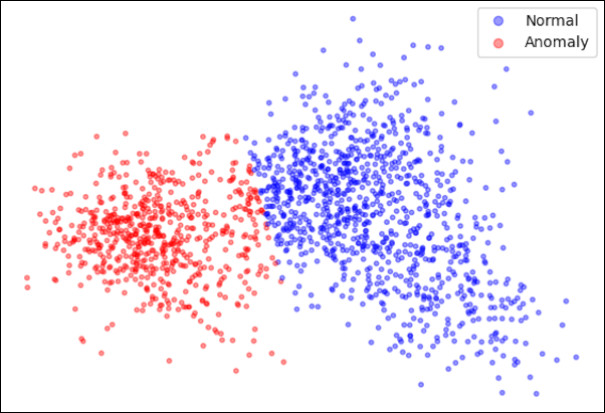} 
        \caption{Clustering: VAE with $\beta=0.01$}
    \end{subfigure}
    
    \caption{Impact of the dimensionality (left) and $\beta=0.01$ (right) on clustering the latent space of a VAE. Adapted from~\cite{Klaus_Anomaly_2022_BA}.}
    \label{fig:ls}
\end{figure}

To classify an image as normal or anomalous during evaluation, K-Means clustering of the latent space of the trained VAE is performed. We used PCA to visualize the distribution of inputs in the latent space. Our experiments have shown that the larger size of the latent space improves the reconstruction strength of the VAE, as shown above, and the clustering in the latent space. A large latent space size $512\times4\times4$ led to better results than small ones like $64\times4\times4$ (see Figure~\ref{fig:ls}). 

A quantitative analysis of cluster assignments for different $\beta$ values, as shown in Table~\ref{tab:fpr_tpr}, has revealed that smaller $\beta$ values lead to lower false positive rates. On the right side of Figure~\ref{fig:ls}, it can be seen that for $\beta=0.01$, the proposed approach can detect most anomalous data, i.e., data points corresponding to three anomaly datasets. 
Adding the previously described cluster loss in Equation~\ref{eq:cluster_loss} did not help to reduce the number of false positives. Furthermore, the distance loss from Equation~\ref{eq:dist_loss} significantly increased the distance between the clusters. However, this latent space structure is unsuitable for cluster separation for normal and anomalous data~\cite{Klaus_Anomaly_2022_BA}.

    

\begin{table}[h]
\centering
\resizebox{0.7\columnwidth}{!}{%
\begin{tabular}{@{}lllll@{}}
\toprule
 $\beta$ & 1 & 0.1 & 0.01 & 0.001 \\ \midrule
 FPR & 0,4332 & 0,3231 & 0,3557 & 0,4065  \\
 TPR & 0,9894 & 0,9681 & 1 & 1  \\ \bottomrule
\end{tabular}%
}
\caption{False and true positive rate for anomaly classification using a VAE with latent space for different $\beta$ values.}
\label{tab:fpr_tpr}
\end{table}



    

\section{Conclusion}
\label{sec:conclusion}

In this work, we have presented an approach to detect image samples containing anomalies based on the latent space of a Variational Autoencoder. The latent space was conditioned in a way to create individual clusters for those categories, which allowed for the detection of anomalies during inference. We could show, that our model is even able to detect small anomalies from datasets without a domain shift compared to the training data. However, similar to other anomaly detection approaches~\cite{du_vos_2022}, our method still produces many false positives. We have performed experiments with different components, such as a distance loss, a cluster loss, or an additional discrepancy map as the input, evaluating their impact on the performance of the model. While high false-positive rates are not suitable for production systems, our approach can be utilized for an active learning system, where a human oracle can choose relevant frames from a pre-selection, based on the detection results from our method.

\section{Acknowledgment}
\label{sec:acknowledgment}
This work results partly from the project KI Data Tooling (19A20001J), funded by the German Federal Ministry for Economic Affairs and Climate Action (BMWK).

{\small
\bibliographystyle{IEEEtran}
\bibliography{references}
}

\end{document}